          [2001/04/25 1.1 (PWD)]
\documentclass[a4paper,twocolumn]{esapub2005} % European paper
\usepackage{times}
\usepackage{url}
\usepackage{amsmath,graphicx}
\usepackage{algorithm,algorithmic}

\usepackage[style=numeric,giveninits=true]{biblatex}
\title{Real-Time Instrument Planning and Perception for Novel Measurements of Dynamic Phenomena}
\author{Itai Zilberstein\thanks{Current affiliation: School of Computer Science, Carnegie Mellon University. Correspondence Author: Steve Chien, steve.a.chien@jpl.nasa.gov.}}
\author{Alberto Candela}
\author{Steve Chien}
\affil{Jet Propulsion Laboratory, California Institute of Technology, Pasadena, CA, USA}

\begin{document}

\keywords{Remote sensing, volcanic plumes, computer vision}

\maketitle

\begin{abstract}
Advancements in onboard computing mean remote sensing agents can employ state-of-the-art computer vision and machine learning at the edge. These capabilities can be leveraged to unlock new rare, transient, and pinpoint measurements of dynamic science phenomena. In this paper, we present an automated workflow that synthesizes the detection of these dynamic events in look-ahead satellite imagery with autonomous trajectory planning for a follow-up high-resolution sensor to obtain pinpoint measurements. We apply this workflow to the use case of observing volcanic plumes. We analyze classification approaches including traditional machine learning algorithms and convolutional neural networks. We present several trajectory planning algorithms that track the morphological features of a plume and integrate these algorithms with the classifiers. We show through simulation an order of magnitude increase in the utility return of the high-resolution instrument compared to baselines while maintaining efficient runtimes. 
\end{abstract}

\section{Introduction}

Significant advances in computer vision and machine learning can enable unique science measurements by allowing for the pointing and reconfiguration of instruments to acquire the most valuable science data.  Such techniques apply to static features and the AEGIS system on Mars rovers is an example.  However, even greater benefits can be achieved by targeting rare, dynamic features such as plumes, deep convective ice storms, or atmospheric events.  Many such events would be hazards to human spaceflight (such as coronal mass ejections) and therefore these techniques can support human exploration as well.

To date, space mission instruments are directed open-loop by ground operations teams days to weeks in advance.  A notable exception is the AEGIS targeting systems used on the MER, MSL, and M2020 rovers on Mars that target static geological features \cite{aegis}. This effort seeks to advance techniques to allow moving spacecraft to target dynamic science features, such as volcanic plumes, and execute novel measurements in real-time. Precise measurements of plumes can supply key data that propels the understanding of processes \cite{plumevariables}.  Ash from volcanic plumes also poses an aviation hazard, and detecting the volume and height of plumes from space can provide timely warnings \cite{measuringplumes}. For ground sensors, leveraging computer vision has been shown to effectively detect plumes \cite{groundplumesegmentation}. We aim to deploy similar systems to remote sensing agents to unlock new measurements of plumes. 

\begin{figure}[t]
    \centering
    \includegraphics[width=\linewidth]{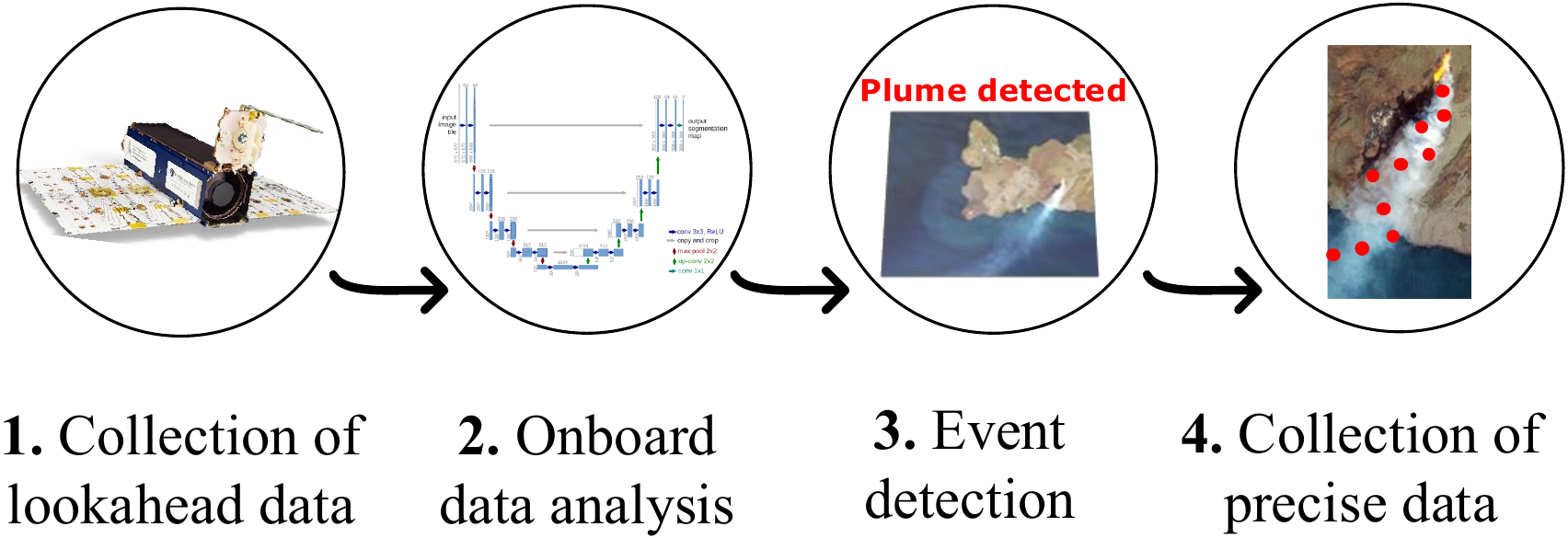}
    \caption{Workflow for dynamic targeting of volcanic plumes. A satellite captures a look-ahead image of a volcano. This data is analyzed onboard to detect plumes. Following detection, precise follow-up measurements of the plume are collected (illustrated with red dots in step 4). \textit{Imagery from Planet Labs (step 1), Ronneberger et al. \cite{unet} (step 2), ESA (steps 3 \& 4).}}
    \label{fig:workflow}
\end{figure}

We focus on the version of the problem for a satellite in Low Earth Orbit (LEO). Remote sensing with LEO satellites has many science applications and LEO satellites have produced lots of data that can be leveraged. Our approach consists of two parts: first, wide field of view (WFOV) imagery is analyzed in real-time to enable subsequent measurement. This analysis leverages computer vision to identify plumes within raw data. Second, using the classification as a heuristic,  we plan and execute follow-up measurements with a narrow field of view (NFOV) instrument. Figure \ref{fig:workflow} depicts this workflow for measuring volcanic plumes. This workflow generalizes to other remote sensing applications including measuring extraterrestrial plumes, wildfires, and harmful algal blooms. We intend to explore these applications and other dynamic phenomena in future work. 

\textit{Dynamic Targeting} (DT), which is the paradigm of utilizing a WFOV instrument to target a NFOV instrument, has been researched previously in simulation \cite{breitfeld-isairas-2024,dt-jais,dt-igarss,dt-planrob-2024}, and flight demonstrations are on-going for LEO use-cases \cite{dt-astra-2025,dt-spaceops-2025,nos-igarss,nos-isairas}. DT has been leveraged for cloud avoidance and generated a $+80\%$ science return for the JAXA TANSO-FTS-2 instrument on GOSAT-2 \cite{suto2021thermal}. According to another study, only $7{-}12\%$ of $CO_2$ or $CH_4$ observations from LEO satellites yield successful measurements with the remaining $\sim90\%$ primarily blocked by clouds \cite{nassar2023intelligent}. DT has been shown to improve science yield when subjected to operational and power constraints \cite{dt-jais,nassar2023intelligent,suto2021thermal}. 

However, previous research has primarily examined the planning portion of DT for the NFOV instrument, and no study has applied DT to the volcanic plume use case. For LEO satellites, executing DT onboard requires efficient real-time planning and perception to ensure that the satellite does not pass over the target before taking observations. This time constraint is typically on the order of $10$s to $100$s of seconds \cite{nos-isairas}. In this paper, we synthesize the trajectory planning of the NFOV sensor with the perception from the WFOV sensor and demonstrate huge increases in the science utility of measurements for the volcanic plume use-case. We evaluate several classification approaches including convolutional neural networks and integrate these models with trajectory planning algorithms in an end-to-end workflow that would operate onboard. This integration leverages morphological operations to extract and denoise polygons from the classification for smooth sensor trajectory planning. 

\section{Planning and Perception}

In this section, we detail the automated perception and planning that generate the trajectory of the NFOV sensor from the data supplied by the WFOV sensor. Algorithm \ref{alg:planning} outlines this onboard process. Using a classifier $\mathcal{C}$, we obtain a segmentation of the plume in the WFOV image. Working with real data means that these masks are noisy. Leveraging computer vision shape analysis, we denoise the classified masks. We then use the trajectory planning algorithm $\lambda$ with the denoised plume mask to compute a trajectory for the NFOV sensor.  We further detail each of these steps. 

\begin{algorithm}[b]
 \caption{Dynamic Plume Planning}
 \label{alg:planning}
 \begin{algorithmic}[1]
 \renewcommand{\algorithmicrequire}{\textbf{Input:}}
 \renewcommand{\algorithmicensure}{\textbf{Output:}}
    \REQUIRE WFOV image $I$, classifier $\mathcal{C}$, trajectory algorithm $\lambda$
    \ENSURE  NFOV trajectory $\pi$ \\
    \STATE $P = \mathcal{C}(I)$ \\
    \STATE $P' = \textsc{Denoise}(P)$ \\
    \STATE $\pi = \lambda(P')$ \\
    \RETURN $\pi$ 
\end{algorithmic} 
\end{algorithm}

\begin{figure}[t]
    \centering
    \includegraphics[width=0.49\linewidth]{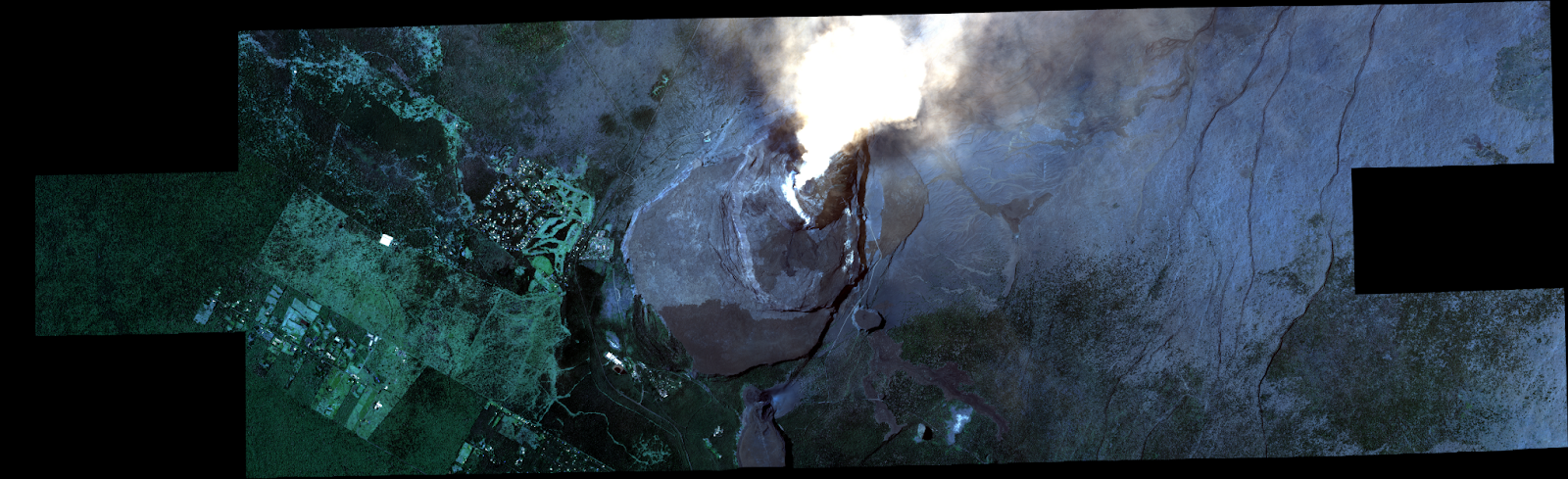}
    \includegraphics[width=0.49\linewidth]{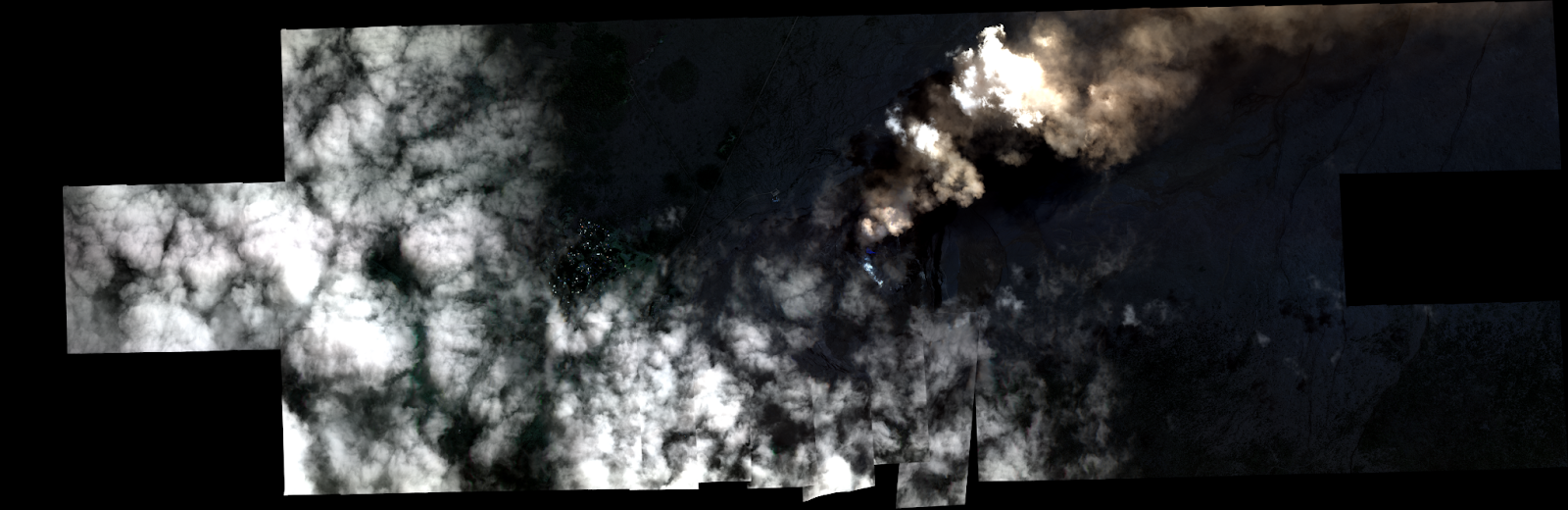}
    \includegraphics[width=0.49\linewidth]{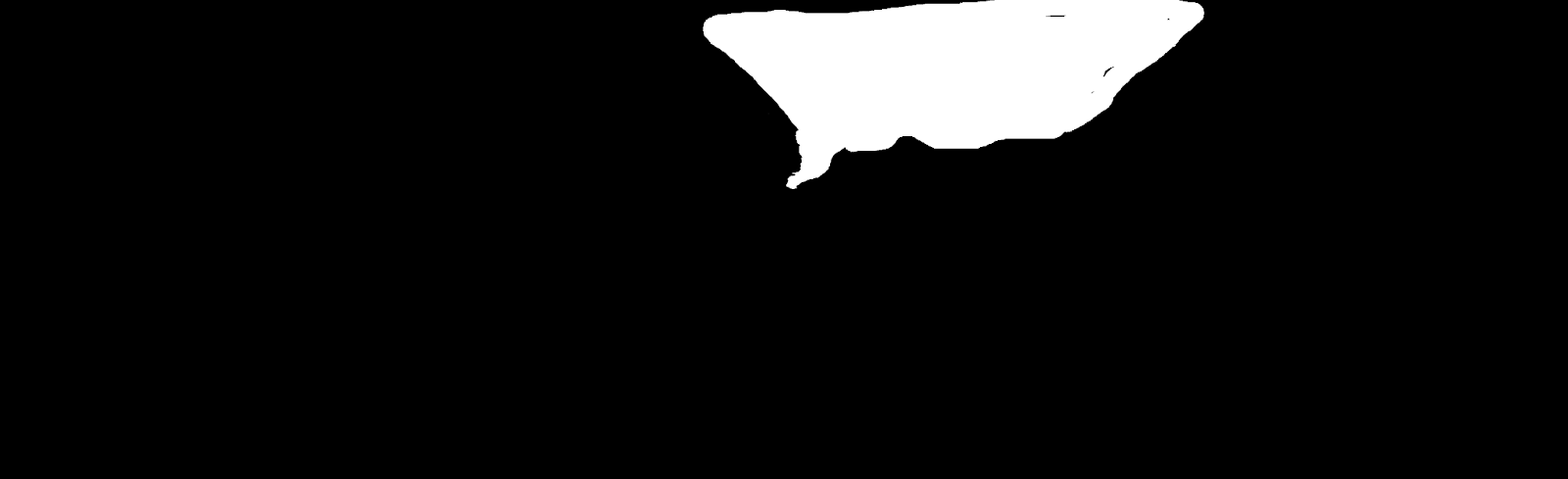}
    \includegraphics[width=0.49\linewidth]{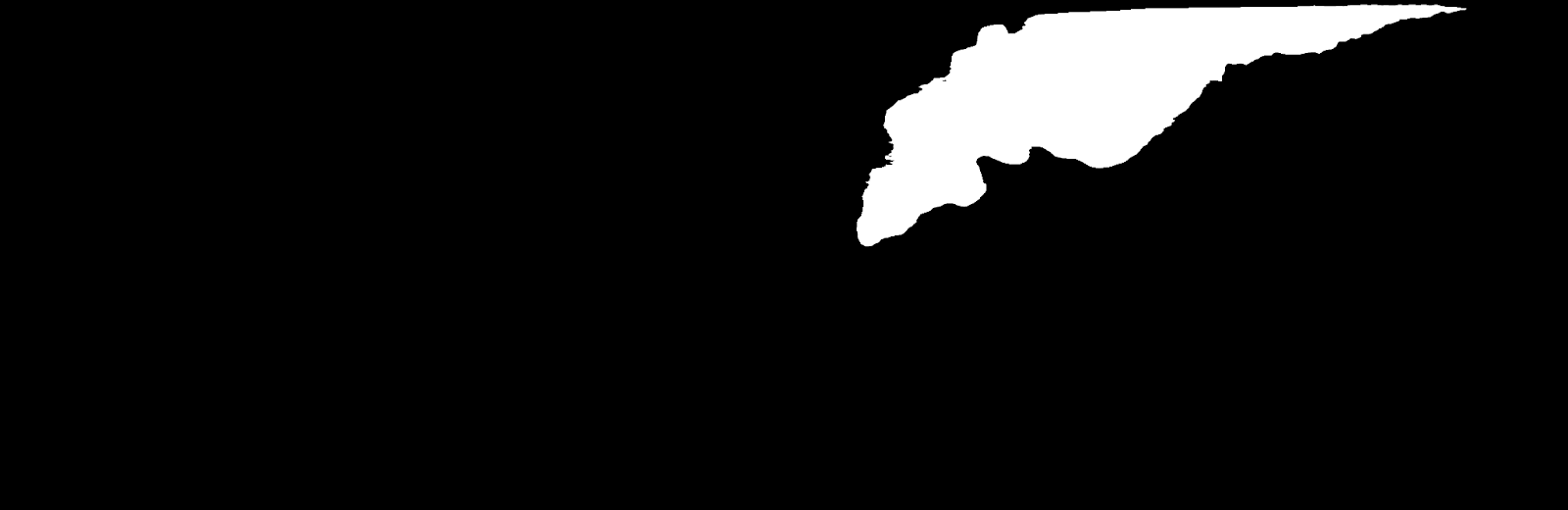}
    \caption{Top: Planet SkySat images of volcanic plumes from the Kilauea volcano. Bottom: Corresponding plume hand-labels. \textit{Imagery from Planet Labs.}}
    \label{fig:dataset}
\end{figure} 

\subsection{Datasets}

The dataset consists of Planet SkySat imagery of volcanoes and their plumes at $0.5$m Ground Sampling Distance (GSD). We downsample these images to simulate WFOV data and use the original high-resolution data products to extract the utility of the NFOV measurements. Skysat images have 4 spectral bands: red, green, blue, and near-infrared (NIR). We utilize $38$ scenes of volcanic plumes that are hand-labeled by experts using LabelBox. This dataset is used for the plume classification and trajectory planning algorithms. We use $21$ of these plumes for training of the classifiers. Figure \ref{fig:dataset} shows some examples of the images that comprise the dataset.

%Figure 1: 
\subsection{Plume Segmentation}

We explore two different approaches to the automatic detection of plumes from satellite imagery: single-pixel classification and image segmentation using deep learning.

The single-pixel classification analyzes each pixel individually and decides if it corresponds to a plume only using spectral information. We include one baseline classifier that performs band thresholding derived from cloud detection to identify plumes \cite{hot}. In the visible wavelengths, clouds and plumes appear very similar, therefore we include the near-infrared band in the band thresholding to account for this distinction. We evaluate the following five single-pixel classifiers, which are developed using scikit-learn \cite{scikit}: Naive Bayes (Gaussian), Logistic Regression, Decision Tree, Random Forest, and Multilayer Perceptron. While Naive Bayes assumes that all spectral bands are independent, the other classifiers do not work under this simplifying assumption. 

We use convolutional neural networks to perform a more complex spatial analysis. We evaluate two U-Net architectures: UNET-Uavsar and UNET-Xception \cite{flood,unet}. These have been shown to perform high-quality image segmentation, even while using small training sets and few model layers, and have been demonstrated onboard several LEO satellites in 2024 \cite{nos-isairas}. To reduce model size and input consistency, images are tiled into square patches. We train the models for $50$ epochs optimizing the sparse categorical cross-entropy loss with the Adam optimizer with a learning rate of $10^{-3}$.

All the models are engineered to execute in a simulated onboard environment in real-time. The models take as input the raw, four-channel WFOV satellite image and output a binary pixel mask that classifies the plume.

\subsection{Denoising}

\begin{algorithm}[t]
 \caption{Denoise}
 \label{alg:denoise}
 \begin{algorithmic}[1]
 \renewcommand{\algorithmicrequire}{\textbf{Input:}}
 \renewcommand{\algorithmicensure}{\textbf{Output:}}
    \REQUIRE Binary plume mask $P$ \\
    \ENSURE Denoised plume mask $P'$\\ 
    \STATE $\text{polygons} = \textsc{getContours}(P)$ \\
    \STATE $\text{mergedPolygons} = \textsc{erodeAndDilate}
    (\text{polygons})$ \\
    \STATE largePolygons $= \textsc{filterByArea}(\text{mergedPolygons})$ \\
    \STATE $P' = \textsc{reconstructMask}(\text{largePolygons})$ \\
    \RETURN $P'$ 
\end{algorithmic} 
\end{algorithm}

Due to noise in the plume segmentation, we leverage morphological operations to denoise these outputs prior to planning \cite{jamil2008noise}. Using OpenCV, we obtain the contours of all polygons within the binary mask \cite{opencv}. We then attempt to merge all disjoint polygons by repeatedly eroding and dilating with a $3 \times 3$ kernel \cite{efficientdilation,erosion}. 
After a certain number of iterations, if polygons cannot be merged, we treat them as separate plumes. We then filter out all polygons that have an area below a threshold. This operation is motivated by the domain knowledge that plumes are typically large enclosed shapes. Finally, we reconstruct the mask based on the final polygons. Algorithm \ref{alg:denoise} outlines these steps.

\subsection{Trajectory Planning}

\begin{figure}[t]
    \centering
    \includegraphics[width=\linewidth]{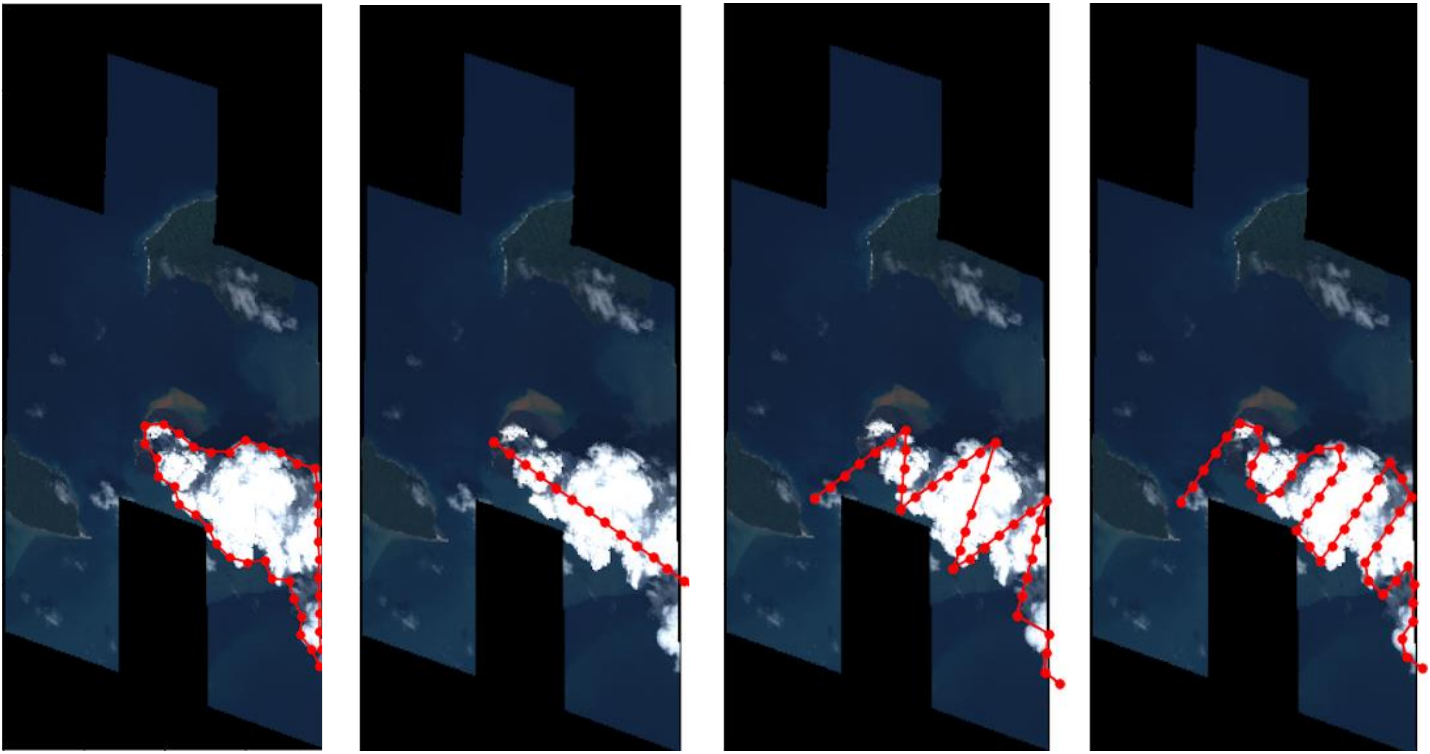}
    \caption{Paths from trajectory algorithms 1, 2, 3, and 4 (left to right) over the Krakatoa Volcano. \textit{Imagery from Planet Labs.}}
    \label{fig:trajectories}
\end{figure}

Using the classified plume as a heuristic, we can intelligently plan the trajectory of the NFOV sensor to obtain high-resolution measurements. The trajectory planning algorithms take as input a static segmentation and employ various policies to plan the trajectory. These algorithms do not perform utility maximization since the actual science return is typically derived from hidden variables. Instead, these algorithms compute geometric trajectories of interest that would capture some of these hidden variables.  We have developed 4 algorithms that enact different trajectories. These algorithms are explained below and shown visually in Figure \ref{fig:trajectories}

\begin{enumerate}
    \item \textit{Trace outline}: this algorithm observes points of interest on the boundary of the plume by tracing the edge points of each disjoint polygon. 
    \item \textit{Track center}: this algorithm efficiently samples points on the edge of and within the plume by identifying and observing along the major axis of each polygon within the plume.
    \item \textit{Diagonal Transect}: this algorithm performs angled transects across the major axes of the plumes. This algorithm obtains a high quantity of diverse plume measurements.
    \item \textit{Lawnmower Transect}: this algorithm performs perpendicular transects across the major axes of the plumes. Like above, this algorithm obtains a high quantity of diverse plume measurements.
\end{enumerate}

We developed custom implementations of transecting algorithms, however, similar methods have been used before for point sampling \cite{transect}. Several of the algorithms compute the major axis of the plume by regressing a line through the contour of each polygon. These algorithms make the simplifying assumption that plumes are straight, which is typical. However, there are cases where plumes curve, and in future work, we can explore computing a piece-wise major axis or fitting a higher degree function such as a spline to the plume. Curves in a plume will primarily affect the \textit{Track Center} algorithm. The transecting algorithms project each cross transect opposite the previous direction and will seek the furthest point on the plume boundary that intersects that line which will drift with a plume's curvature. For the transecting algorithms, the complication with curves is that the transects may not be perpendicular cross-sections across the plume. 

All of the algorithms are parameterized by a step size which defines the number of pixels between measurements along the path. This step size is dynamically computed to be 1\% of the x-axis of the input scene. The transecting algorithms have a parameterized width, which is the distance in pixels between the starting points of consecutive cross-transects. This value is fixed to be twice the step size. In practice, these parameters would be set based on the constraints of the NFOV instrument. We evaluated several different instantiations of these parameters and note that changing these values simply scales the metrics across algorithms and does not impact the relative results of classifiers and algorithms. %A larger step size results in lower utility and a smaller step size results in higher utility. This behavior is what we expect as a lower step size would result in more measurements and vice-versa. Changing this value does not impact the relative results of classifiers and algorithms. 

\section{Results}

All experiments are executed on a MacBook Pro 16 laptop with an M2 Max processor and 64 GB of RAM. This level of computing power is comparable to the next generation of LEO satellites \cite{cs6}. The makespan execution time of the simulations is several seconds.

\subsection{Plume Classification}

\begin{table}[t]
    \label{tab:classification}
    \centering
    \resizebox{\linewidth}{!}{
        \begin{tabular}{c||c|c|c|c}
            Classifier & Accuracy & Precision & Recall & IoU  \\
            \hline \hline
            Band Threshold & $0.806$ &  $0.154$ & $0.577$ & $0.138$ \\
            + Denoising & $0.765$ &  $0.132$ & $0.605$ & $0.122$ \\
            \hline
            Naive Bayes & 0.719 &  0.580 & 0.641 & 0.438 \\
            + Denoising & 0.728 &  0.589 & 0.676 & 0.459 \\
            \hline
            Logistic Regression & 0.794 & 0.837 & 0.490 & 0.448 \\
            + Denoising & 0.776 & 0.726 & 0.552 & 0.457 \\
            \hline
            Decision Tree & 0.839 & 0.920 & 0.580 & 0.552 \\
            + Denoising & 0.873 & 0.927 & 0.681 & 0.646 \\
            \hline
            Random Forest & 0.870 & 0.915 & 0.684 & 0.643 \\
            + Denoising & 0.920 & 0.906 & 0.856 & 0.786 \\
            \hline
            Multilayer Perceptron & 0.766 &  0.767 & 0.454 & 0.399 \\
            + Denoising & 0.770 & 0.762 & 0.473 & 0.412 \\
            \hline
            UNET-Xception & 0.926 &  0.926 & \textbf{1.000} & 0.926 \\
            + Denoising  & 0.926  & 0.926 & \textbf{1.000} & 0.926 \\
            \hline
            UNET-Uavsar & 0.901 & \textbf{0.931} & 0.965 & 0.900 \\
            + Denoising & \textbf{0.927} & 0.927 & 0.999 &  \textbf{0.927} \\
        \end{tabular}
    }
    \caption{Performance of the plume classifiers as evaluated with 5-fold cross validation. The precision, recall, and intersection over union (IoU) is shown for the  plume class.}
\end{table}

We begin by analyzing the performance of the plume classifiers in isolation from the planning algorithms. We report the classification accuracy along with the precision, recall, and intersection over union (IoU) of the positive class (plume) evaluated using 5-fold cross-validation. We also report these metrics after performing denoising on the output segmentation. We mention that the precision, recall, and intersection over union (IoU) for the negative class (non-plume) is above $90\%$ for nearly all of the classifiers due to the large areas of images that do not contain plumes.

\begin{figure}[t]
    \centering
    \includegraphics[width=\linewidth]{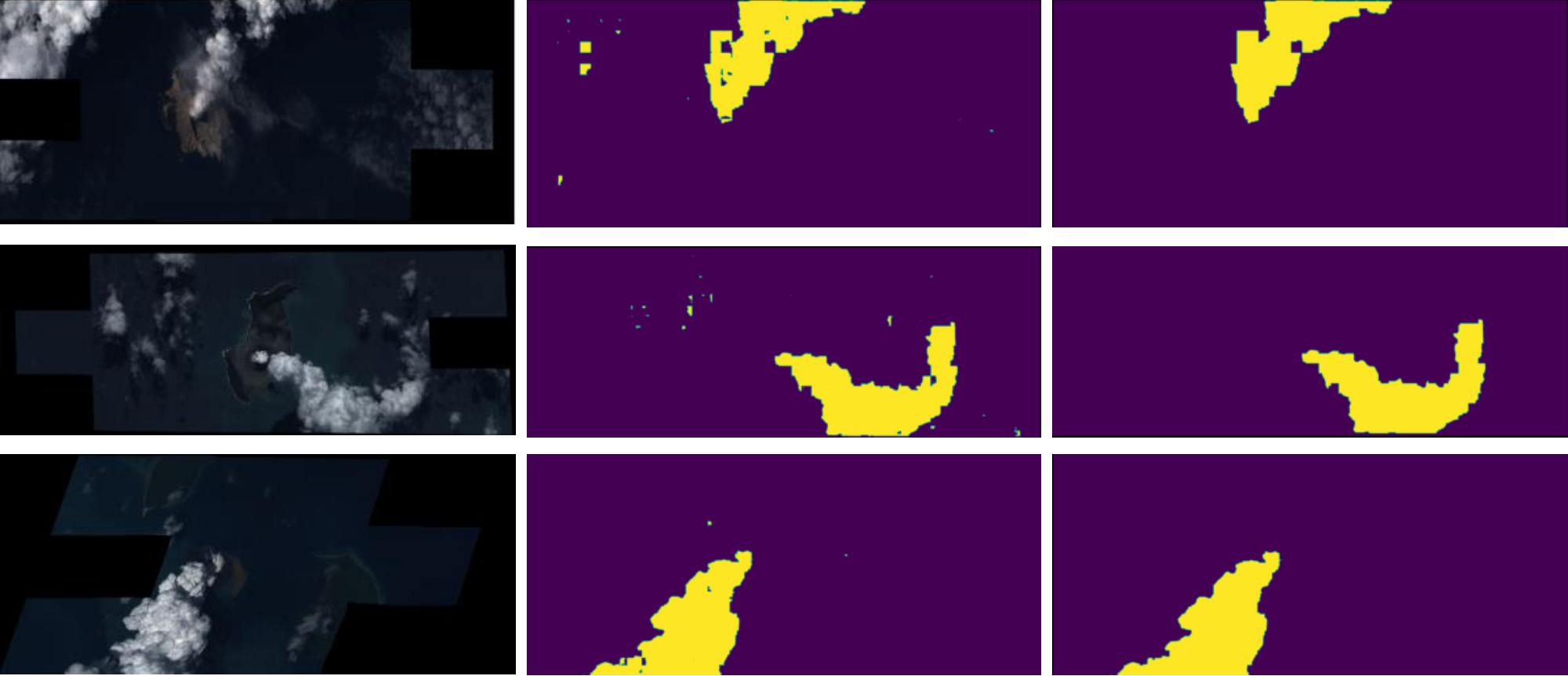}
    \caption{Examples of plume classification results using the UNET-Xception classifier. On the left is the image, the middle is the output from the classifier, and the right is the output after performing denoising. \textit{Imagery from Planet Labs}.}  %These examples correspond to challenging scenarios due to cloud presence and other conditions. 
    \label{fig:rf-classification}
\end{figure}

The classification accuracy results of each classifier are reported in Table \ref{tab:classification}. Naive Bayes has a poor performance likely due to its simplifying assumptions, while the other classifiers that do not work under this simplifying assumption achieve a higher classification accuracy. The UNETs have the best performance across metrics with a classification accuracy of over $92\%$. We notice that denoising generally improves the classification across algorithms, supporting the efficacy of this procedure. Denoising almost always improves the accuracy, recall, and IoU, with a more balanced effect on the precision. Figure \ref{fig:rf-classification} highlights examples of denoised outputs for the UNET-Xception classifier.

\subsection{Planning and Perception}

\begin{table*}[t]
    \label{tab:e2e}
    \centering
    \small
    \begin{tabular}{c|c||c|c|c|c|c}
        Trajectory Algorithm & Classifier & Pixels Observed & Ratio Plume & Intensity & Gradient & Runtime (s)  \\
        \hline\hline
        Straight Nadir & N/A  & 104.000 & 0.081 & 0.003 & 0.005 & 0.001 \\
        \hline
        Naive Transect & N/A & 416.381 & 0.113 & 0.004 & 0.009 & 0.027 \\
        \hline
        Trace Outline
        & Band Threshold & 127.586 & 0.123 & 0.004 & 0.010 & 9.520 \\
        & Naive Bayes & 204.457 & 0.059 & 0.001 & 0.004 & 10.748 \\
        & Logistic Regression & 175.381 & 0.684 & 0.336 & 0.046 & 7.329 \\
        & Decision Tree & 200.429 & 0.661 & 0.304 & 0.046 & \textbf{7.310} \\
        & Random Forest & 316.048 & 0.520 & 0.202 & 0.037 & 8.706 \\
        & Multilayer Perceptron & 100.524 & 0.726 & \textbf{0.441} & 0.048 & 8.863 \\
        & UNET-Xception & 239.048 & 0.687 & 0.210 & \textbf{0.057} & 10.109 \\
        & UNET-Uavsar  & 172.381 & \textbf{0.813} & 0.299 & 0.047 & 8.589 \\
        \hline
        Track Center
        & Band Threshold  & 44.033 & 0.103 & 0.004 & 0.006 & 9.538 \\
        & Naive Bayes & 70.100 & 0.055 & 0.002 & 0.003 & 10.784 \\
        & Logistic Regression & 67.810 & 0.599 & 0.290 & 0.036 & 7.339 \\
        & Decision Tree & 80.190 & 0.555 & 0.256 & 0.032 & \textbf{7.297}\\
        & Random Forest & 134.952 & 0.405 & 0.164 & 0.022 & 8.677 \\
        & Multilayer Perceptron & 42.667 & \textbf{0.691} & \textbf{0.399} & \textbf{0.044} & 8.873 \\
        & UNET-Xception & 89.619 & 0.624 & 0.232 & 0.032 & 10.129 \\
        &  UNET-Uavsar  & 63.667 & 0.687 & 0.280 & 0.038 & 8.611 \\
        \hline
        Diagonal Transect
        & Band Threshold  & 696.043 & 0.114 & 0.004 & 0.006 & 9.571 \\
        & Naive Bayes & 1259.300 & 0.085 & 0.003 & 0.004 & 11.110 \\
        & Logistic Regression & 451.905 & 0.626 & 0.323 & 0.038 & 7.259 \\
        & Decision Tree & 548.429 & 0.610 & 0.302 & 0.037 & \textbf{7.213} \\
        & Random Forest & 867.905 & 0.557 & 0.252 & 0.032 & 8.533 \\
        & Multilayer Perceptron & 222.571 & 0.694 & \textbf{0.414} & \textbf{0.044} & 8.869 \\
        & UNET-Xception & 821.571 & 0.784 & 0.303 & 0.041 & 9.978 \\
        &  UNET-Uavsar & 458.381 & \textbf{0.844} & 0.347 & 0.043 & 8.515 \\
        \hline
        Lawnmower Transect
        & Band Threshold  & 767.110 & 0.120 & 0.005 & 0.007 & 10.802 \\
        & Naive Bayes & 1374.124 & 0.087 & 0.003 & 0.004 & 12.972 \\
        & Logistic Regression & 567.905 & 0.656 & 0.346 & 0.042 & \textbf{7.400} \\ 
        & Decision Tree & 662.571 & 0.642 & 0.324 & 0.040 & 7.407 \\
        & Random Forest & 1012.905 & 0.565 & 0.260 & 0.035 & 8.829 \\
        & Multilayer Perceptron & 269.286 & 0.724 & \textbf{0.438} & \textbf{0.048} & 8.860 \\
        & UNET-Xception & 992.952 & 0.818 & 0.315 & 0.044 & 10.101 \\
        &  UNET-Uavsar  & 553.524 & \textbf{0.873} & 0.358 & 0.043 & 8.651 \\
        
    \end{tabular}
    \caption{Performance of the perception and planning algorithms. We report the average number of pixels observed with the NFOV sensor, the fraction of these pixels that are within the plume, the average normalized intensity of the observations, the average normalized Sobel gradient of the observations, and the runtime which is reported in seconds.}
\end{table*}

Table \ref{tab:e2e} depicts the results of the synthesized planning and perception. For each trajectory algorithm, we run the simulation with each classifier. We input a raw satellite image which gets classified, denoised, and inputted into the trajectory planning. The output of the simulation is the path of observations for the NFOV sensor.  We simulate a NFOV sensor that can observe a single pixel at a time. We report the average pixels observed by the NFOV sensor, the average fraction of these pixels within the plume, and the average make-span runtime in seconds. We also include two utility functions that capture the potential science returns of the NFOV observations that are unknown during planning.  The intensity utility is derived from overlaying the plume label on the blue band of the satellite imagery and performing normalization. This function captures the simulated intensity of a plume pixel. The gradient function is derived from the intensity by computing the normalized Sobel gradient using a $5 \times 5$ kernel. This function captures the interest in volatile features of the plume including the diffusion at the edges.  

We evaluate two baseline trajectory algorithms that do not rely on look-ahead information. The first, \textit{Straight Naidr} observes along the nadir direction of the satellite, which is the standard for many remote sensing missions. \textit{Naive Transect} blindly transects portions of the scene without knowledge of the plume's location.  

The trajectory algorithms offer different capabilities and a comparison depends on the specifics of a mission. \textit{Trace Outline} and \textit{Track Center} offer efficient utility with few observations. \textit{Trace Outline} prioritizes the high gradient observations at the edge of the plume.  Transecting achieves higher plume ratios and intensity, but at the cost of many observations. 

The results show an enormous return in the utility of observations when synthesizing planning with high-quality classifiers. The best classifiers, such as the UNETs, achieve an order of magnitude increase in the ratio of plume observed and intensity utility and two orders of magnitude increase in the gradient utility compared to the baseline trajectory algorithms and the Band Threshold and Naive Bayes classifiers. Over $87\%$ of pixels observed are within the plume for the UNET-Uavsar model compared to $12\%$ for the best baseline. 

We generally observe that the better the classifier, the better the utility of the trajectories. However, the Multilayer Perceptron (MLP) has high utility despite low classification performance. We find that the MLP under-classifies the plume, which is consistent with high precision and low recall. The MLP observes fewer pixels, yet a higher ratio of these are within the plume. Comparing this to classifiers that over-classify, we see that it is beneficial to be conservative.  The intensity and gradient utility are also unknown to the planning algorithms and the utility varies across the plume. The MLP identifies a small, yet high utility region of the plume. 

The runtimes of the workflow are below $13$ seconds. This level of efficiency meets the $10$ to $100$s timing constraints required to perform dynamic targeting. Further experiments on flight hardware are required to confirm these results.

\section{Conclusion}

Deploying image processing onboard satellites can revolutionize remote sensing. By identifying points of interest, an autonomous agent can dynamically target its instruments to obtain the most precise measurements.  We have shown a huge increase in the utility of measurements of volcanic plumes when using dynamic targeting with high-quality classifiers. 

In future work, we aim to build a large volcanic plume dataset for robustly benchmarking these classifiers. Instead of segmenting pixels within the plume, we can train classifiers that predict a continuous utility and perform utility maximization planning. In addition, we are looking to obtain video data of volcanic eruptions and plumes that can be used for time series analysis. Video analysis enables the estimation of plume rise velocities and top heights  \cite{video}. 

Dynamic targeting will also feature in NASA's upcoming \textit{Federated Autonomous Measurement} (FAME) demonstration \cite{fame-spaceops-2025}. FAME will showcase coordinated measurements from a large, multi-agent, federated observing system composed of many spacecraft \cite{zilberstein-spaceops-2025,zilberstein-jair-2025}. Targeting volcanic plumes is one of multiple applications covered in the FAME demonstration.

 As of July 2025, dynamic targeting has been demonstrated in-orbit on the CogniSAT-6 spacecraft for the cloud avoidance use-case \cite{dt-astra-2025}. We are evaluating spacecraft to fly the technology outlined in this paper, and have already identified satellites to execute the planning portion. However, highly agile sensors are rare, meaning that many spacecraft are not suited for dynamic measurement. We seek to extend this work to other spacecraft and applications including observing wildfire plumes and extraterrestrial plumes on Europa, Enceladus, and comets. Precise measurements of these plumes will offer new insights into their key properties.

\section*{Acknowledgments}
The research was carried out at the Jet Propulsion Laboratory, California Institute of Technology, under a contract with the National Aeronautics and Space Administration (80NM0018D0004). Distributed under 2025 License CC BY-NC-ND 4.0.

\printbibliography

\end{document}